\documentclass{article}
\usepackage[T1]{fontenc}
\usepackage{spconf,amsmath,graphicx,hyperref}
\usepackage{booktabs}
\usepackage{multirow}
\usepackage{amssymb}
\usepackage{graphicx,verbatim}
\usepackage{caption}
\usepackage{adjustbox}

\title{TCATSeg: A Tooth Center-Wise Attention Network for 3D Dental Model Semantic Segmentation}

\name{
\begin{tabular}{c}
Qiang He$^{1,2}$\hspace{0.5em}
Wentian Qu$^{1,2}$\hspace{0.5em}
Jiajia Dai$^{3}$\hspace{0.5em}
Changsong Lei$^{3}$\hspace{0.5em}
Shaofeng Wang$^{4}$\hspace{0.5em}
Feifei Zuo$^{5}$ \\
Yajie Wang$^{5}$\hspace{0.5em}
Yaqian Liang$^{3,\dagger}$\hspace{0.5em}
Xiaoming Deng$^{1,2}$\hspace{0.5em}
Cuixia Ma$^{1,2,\dagger}$\hspace{0.5em}
Yong-Jin Liu$^{3}$\hspace{0.5em}
Hongan Wang$^{1,2}$
\end{tabular}
\thanks{Cuixia Ma and Yaqian Liang are corresponding authors.}
}
  
\address{
$^{1}$Institute of Software, Chinese Academy of Sciences  
$^{2}$University of Chinese Academy of Sciences \\
$^{3}$Department of Computer Science and Technology, Tsinghua University \\
$^{4}$Beijing Stomatological Hospital, Capital Medical University 
$^{5}$LargeV Instrument Corporation, Ltd.}

\begin{document}

\maketitle
\begin{abstract} 
Accurate semantic segmentation of 3D dental models is essential for digital dentistry applications such as orthodontics and dental implants. 
However, due to complex tooth arrangements and similarities in shape among adjacent teeth, existing methods struggle with  accurate segmentation, because they often focus on local geometry while neglecting global contextual information. 
To address this, we propose TCATSeg, a novel framework that combines local geometric features with global semantic context.   We introduce a set of sparse yet  physically meaningful superpoints to capture global semantic relationships and enhance segmentation accuracy.   Additionally, we present a new dataset of 400 dental models, including pre-orthodontic samples, to evaluate the generalization of our method.   Extensive experiments demonstrate that TCATSeg outperforms state-of-the-art approaches.
\end{abstract}


\section{Introduction}
\label{sec:intro}
3D dental models derived from intraoral optical scanners are widely used in digital dentistry,  enabling comprehensive morphological analysis and precision treatment planning in orthodontic care. 
Accurate automatic semantic segmentation of these models is essential for various computer-aided clinical applications, including orthodontics~\cite{fan2024collaborative} and implants~\cite{ren2024high}.

\begin{figure}[h]
    \centering
    \includegraphics[width=\linewidth]{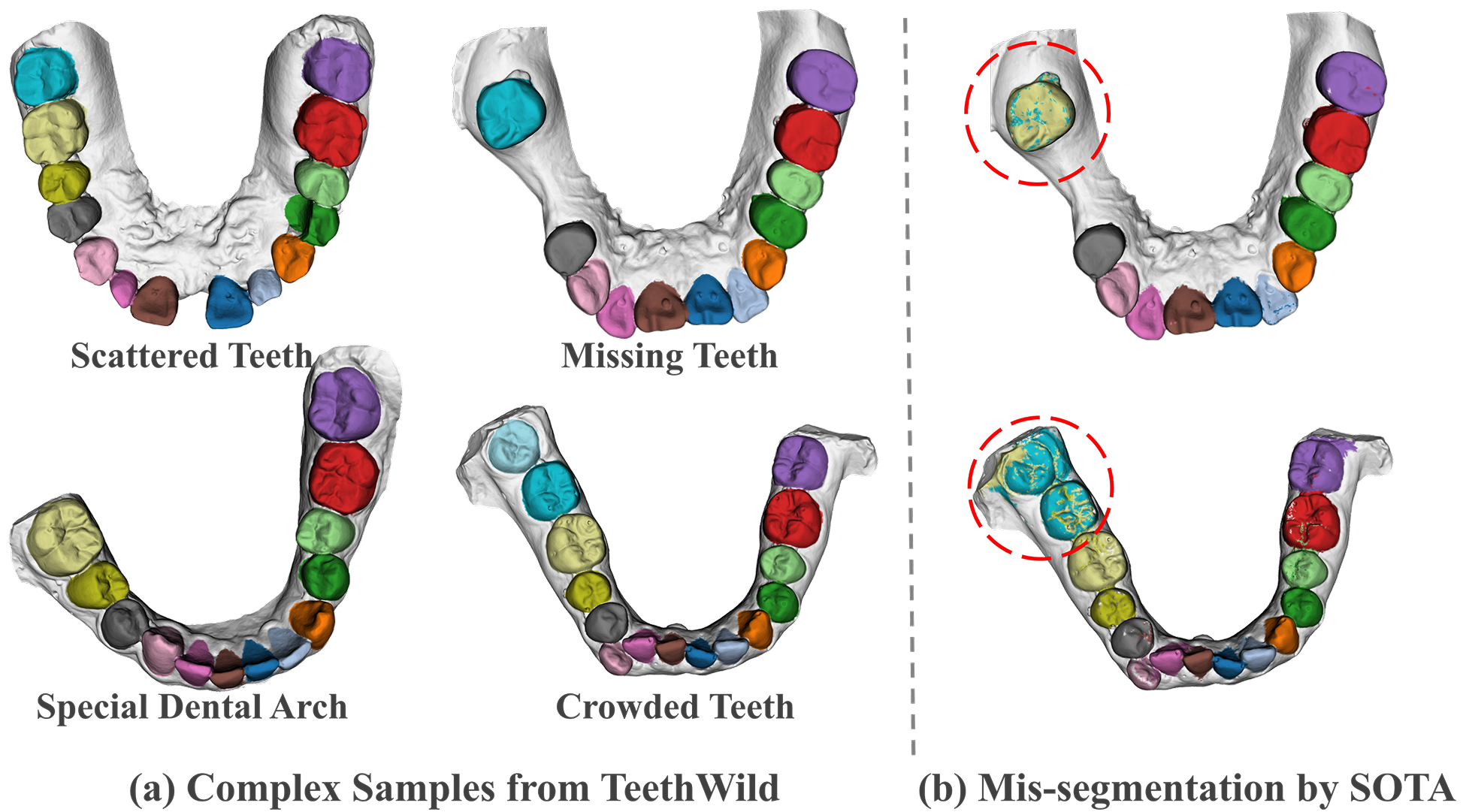}
    \caption{SOTA method~\cite{zhao2021point} struggles on our TeethWild dataset.}
    \label{fig:complex_sample}
\end{figure}

Recently, many point cloud-based 3D dental model segmentation methods have been proposed, including PointNet-based methods~\cite{cui2021tsegnet,lian2019meshsnet}, graph neural networks~\cite{liu2023grab,wu2022two,zhang2021tsgcnet,zheng2022teethgnn},
and attention mechanisms~\cite{yan2020pointasnl,choe2022pointmixer,lin2023dbganet,mazur2021cloud,park2023self,sun2023superpoint,xiong2023tsegformer,zhao2021point,zhuang2023robust}. 
However, due to the unique characteristics of 3D dental models, several challenges remain. Adjacent teeth, like the first and second premolars, often have similar shapes.
Furthermore, 3D dental models derived from real clinical patients often contain complex tooth arrangements, including missing, crowded, or scattered teeth, as well as special dental arch (as shown in Fig.~\ref{fig:complex_sample}).
These challenges hinder methods~\cite{cui2021tsegnet,huang2023lcpformer,lian2019meshsnet,qi2017pointnet++} that rely solely on local geometric features, making accurate segmentation difficult.
To address these issues, a comprehensive approach that integrates both local geometric features and global semantic context is required.
Local geometric features are crucial for delineating the boundaries between adjacent teeth and between teeth and gums, while a global semantic context is essential for distinguishing instances of teeth with similar appearances.
Although several efforts~\cite{lin2023dbganet,xiong2023tsegformer} have attempted to utilize global self-attention to capture global geometric structures, they lack explicit guidance mechanisms, which limit their ability to effectively capture global information.
In contrast, SpoTr~\cite{park2023self} introduces self-positioning points to assist in capturing global context for point cloud understanding tasks. However, dental model exhibits highly similar shapes compared to the normal 3D model, resulting in the self-positioning points being ineffective, as shown in the subsequent experiment (Fig. \ref{fig:hyperpoints}).

\begin{figure*}[!ht]
\centering
\includegraphics[width=0.9\textwidth]{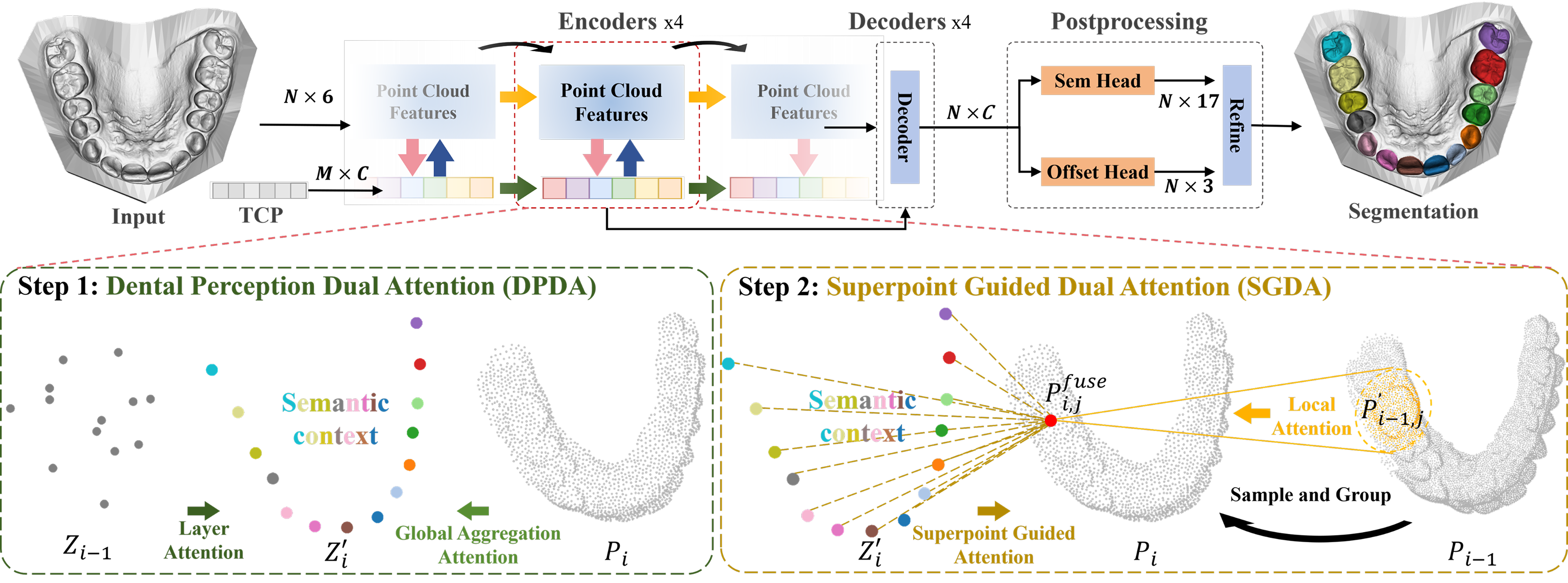} 
\caption{Pipeline of TCATSeg. 
The encoder introduces two novel components: 
DPDA, which enables superpoints  (TCP) to capture semantic context, and SGDA, which integrates local geometric features with global semantic context.
}
\label{fig-over}
\end{figure*}

Inspired by previous studies~\cite{cui2021tsegnet,lin2023dbganet,park2023self}, we observe that the natural arrangement of teeth encodes the contextual relationships between them.
Building on this insight, we propose TCATSeg, a novel framework that leverages sparse superpoints to align feature representations with the arrangement of teeth, thereby capturing global semantic context. To enhance point-wise feature learning, TCATSeg incorporates a dual attention mechanism that integrates local geometric features with global semantic context simultaneously.
Furthermore, we collected a dataset to assess generalization.

To summarize, our main contributions are:
1) We propose TCATSeg, a novel superpoint-guided network for tooth segmentation, which is the first to leverage physically meaningful superpoints to capture the global semantic context of 3D dental models.
2) We design a dual attention module to effectively integrate both local geometric features and global semantic context extracted from superpoints into the point-wise feature learning to improve segmentation accuracy.
3) We construct a new challenging dataset of 400 dental models for cross-domain generalization evaluation, including various pre-orthodontic samples. Extensive experiments on this dataset and benchmarks demonstrate that our TCATSeg outperforms state-of-the-art methods.

\section{Method}
\label{sec:format}
\subsection{Overview}
As shown in Fig.~\ref{fig-over}, we downsample $N=10,000$ points from the input 3D dental model and extract the 3D coordinates and normal vectors as initial features. 
Our architecture comprises four Teeth Center-Wise (TCW) encoders and four skip-connected symmetric decoders, where each TCW encoder downsamples the input point set by a factor of 4, and the corresponding decoder module then upsamples the point features using linear interpolation with a factor of 4.
Subsequently, we utilize two regression heads for point offset and semantic label prediction.
Finally, we adopt a refinement strategy in post-processing to refine the semantic labels for raw-resolution segmentation as previous studies~\cite{ben20233dteethseg,cui2021tsegnet}.
\subsection{Teeth Center-Wise Encoder}
In this section, we detail two key innovations within the proposed Teeth Center-Wise (TCW) encoder: Dental Perception Dual Attention (DPDA) and Superpoint Guided Dual Attention (SGDA). Both modules leverage the Channel-Wise Attention (CWA)~\cite{park2023self,zhao2021point} for learning powerful and flexible representations through adaptive channel-weighting operations.

\vspace{1mm}
\noindent {\bf Channel-Wise Attention.}
The CWA computes distinct channel weights between query and key points. It is formulated as
$
\mathcal{C}(q,k)=Softmax\!\big(W(f_{q}-f_{k},\,x_{q}-x_{k})\big)
$,
where $x_{q},x_{k} \in \mathbb{R}^{3}$ are 3D coordinates  of query and key, $f_{q},f_{k} \in \mathbb{R}^{C}$ are feature vectors, and $W$ denotes the mapping functions.

The updated query set $P_q'=\{X_{q}, F_{q}'\}$ is obtained by cross-attention between the query $P_q$ and the key $P_k = \{X_{k}, F_{k}\}$, expressed as 
$
    P_{q}' = CWA(P_{q}, P_{k}).
$
Each updated feature $f_q' \in F_q'$ is computed as
$
    f_{q}' = \sum\limits_{x_{k},f_{k}} \mathcal{C}(q,k) \odot R({f_{q}, f_{k}})
$
where $x_k \in X_k$, $f_k \in F_k$, $\odot$ is element-wise multiplication, and $R$ is  subtraction function.

\vspace{1mm}
\noindent {\bf Dental Perception Dual Attention.}
We propose using sparse superpoints to capture the global context, illustrated in Fig.\ref{fig-over}.
To learn robust global semantics, we design a dual attention module, comprising Global Aggregation Attention (GA)  and Layer Attention (LayA).  GA aggregates features across spatial and semantic domains via channel-wise attention on the original point cloud, and LayA enforces consistency of superpoint global semantics across all TCW encoder layers. 
As the spatial distribution of tooth centers naturally encodes the contextual relationships between teeth, we constrain the superpoints' spatial arrangement to align with the tooth centers, explicitly guiding them to capture the global context.

Consequently, we define a set of superpoints, termed TCP($M=16$), whose positions approximate tooth centroids and whose features encode the semantic context of the teeth. 

For the $i$-th TCW encoder, the TCP and input point cloud are denoted as $Z_i = \{Y_i, H_i\}$ and $P_{i-1} = \{X_{i-1}, F_{i-1}\}$, where $Y_i$ and $X_{i-1} $ are 3D coordinates, and $H_i$ and $F_{i-1}$ are their corresponding features.
We downsample $P_{i-1}$ via farthest point sampling to obtain $P_{i} = \{ X_{i},F_{i}\}$, and initialize $H_i$ as learnable vectors following ~\cite{dai2017guodongzhang,park2023self}.
TCP positions $Y_{i}$ are interpolated from $P_{i}$ as
$Y_{i} = Softmax(H_{i}F_{i}^T)X_{i}.$
We use TCP as the query in the GA to aggregate the point-wise feature. 
To maintain interlayer consistency of the TCP, we employ LayA between $Z_{i-1}$ and $Z_{i}$. 
The dual attention process is formulated as 
$
H_{i}' = \beta \text{CWA}(Z_{i}, P_{i}) + (1-\beta) \text{CWA}(Z_{i}, Z_{i-1}),
$
where $\beta$ is a learnable parameter that balances global context and inter-layer consistency.
TCP positions are refined via interpolation on $P_i$:
$Y_{i}' = Softmax(H_{i}'F_{i}^T)X_{i}$,
with the refined coordinates supervised by $\mathcal{L}_{tcp}$
to align with ground-truth dental centroids.
The updated TCP can be formulated as $Z_{i}' = \{Y_{i}', H_{i}' \}$.

\vspace{1mm}
\noindent {\bf Superpoint Guided Dual Attention.}
As shown in Fig. \ref{fig-over}, our proposed SGDA framework enhances point-wise feature learning through two complementary mechanisms: Local Attention (LocA) for capturing individual teeth's geometric structures and a Superpoint Guided Attention (SG) for injecting semantic context from superpoints.
Following previous studies~\cite{qi2017pointnet++,park2023self,zhao2021point}, 
we perform the ball query on $P_{i-1}$ to aggregate neighborhood features, resulting in $P_{i-1}'$.
We then implement local attention over the neighborhood of each point, formulated as $ P_{i,j}^{local} = CWA(P_{i,j}, P_{i-1,j}')$, where $P_{i,j}$ denotes the $j$-th point in $P_{i}$ and $P_{i-1,j}'$ corresponds to its local neighborhood.
Simultaneously, to integrate the global semantic context, we compute $P_{i,j}^{global} = CWA(P_{i,j}, Z_{i}')$ using the TCP feature $Z_{i}'$ obtained through DPDA.
The dual-stream features are then adaptively fused with learnable weights: $P_{i,j}^{fuse} = \alpha P_{i,j}^{global} + (1 - \alpha) P_{i,j}^{local}$, where $\alpha$ is a trainable weight.

\subsection{Loss Functions}
\label{sec:pagestyle}

To handle varying tooth counts and unknown correspondence between predicted TCP $p$ and ground-truth centroids $g$, we use 3D Hungarian matching ~\cite{kuhn1955hungarian} to establish optimal pairings, followed by smooth $L_{1}$ loss computation for dentition perception supervision loss $\mathcal{L}_{tcp}$:
\begin{equation}
\mathcal{L}_{tcp} = \sum_{(i,j) \in M(p,g)} L_{1}^{smooth}(p_{i},g_{j})
\end{equation}
where $(i,j)$ are the indices in the matched pairs $M(p,g)$.

Following previous methods~\cite{ben20233dteethseg,cui2021tsegnet}, we employ the cross-entropy loss $\mathcal{L}_{seg}$ for segmentation head and the Chamfer Distance loss $\mathcal{L}_{offset}$  between the predicted offset $\hat{O}$ and the ground truth offset $O$ to constrain the offset head:

\begin{equation}
    \mathcal{L}_{offset} = \frac{1}{|O|}
    \Big(\sum_{o \in O} \min_{ \hat{o} \in \hat{O}} \| o - \hat{o}\|_{2}^{2} + \sum_{\hat{o} \in \hat{O}} \min_{ o \in O} \| \hat{o} - o\|_{2}^{2}
    \Big)
\end{equation}

The total loss can be formulated as
$
\mathcal{L}_{total} = \mathcal{L}_{seg} + \mathcal{L}_{tcp} + \mathcal{L}_{offset}
$.

\vspace{-2mm}
\section{Experiments}

\subsection{Datasets and Experimental Setup}
We evaluated methods on the Teeth3DS dataset~\cite{ben2022teeth3ds} with patient-wise split, ensuring each patient's upper and lower jaws remain in the same partition. We also follow the official protocol of the 3DTeethSeg'22 challenge~\cite{ben20233dteethseg} for fair comparison.
To assess clinical generalization, we introduce TeethWild, a dataset of 400 3D dental models from orthodontic patients at a specialized hospital. As shown in Fig.~\ref{fig:complex_sample}, it includes challenging cases such as abnormal arches, severe crowding, deep overbite/overjet, misalignment, and missing teeth. Annotations were performed by four board-certified orthodontists using Mesh Labeler under the FDI numbering system, with discrepancies resolved via dual validation by a senior clinician. 
For evaluation, we adopt commonly used metrics~\cite{lin2023dbganet}: overall accuracy (OA), dice similarity coefficient (DSC), sensitivity (SEN) and positive predictive value (PPV).
We also adopt the official 3DTeethSeg'22 metrics~\cite{ben20233dteethseg}, including tooth localization accuracy (TLA), 
teeth segmentation accuracy (TSA),
and tooth identification rate (TIR).

\subsection{Comparison with State-of-the-Arts}

\begin{figure}[!t]
\centering

\includegraphics[width=1\linewidth]{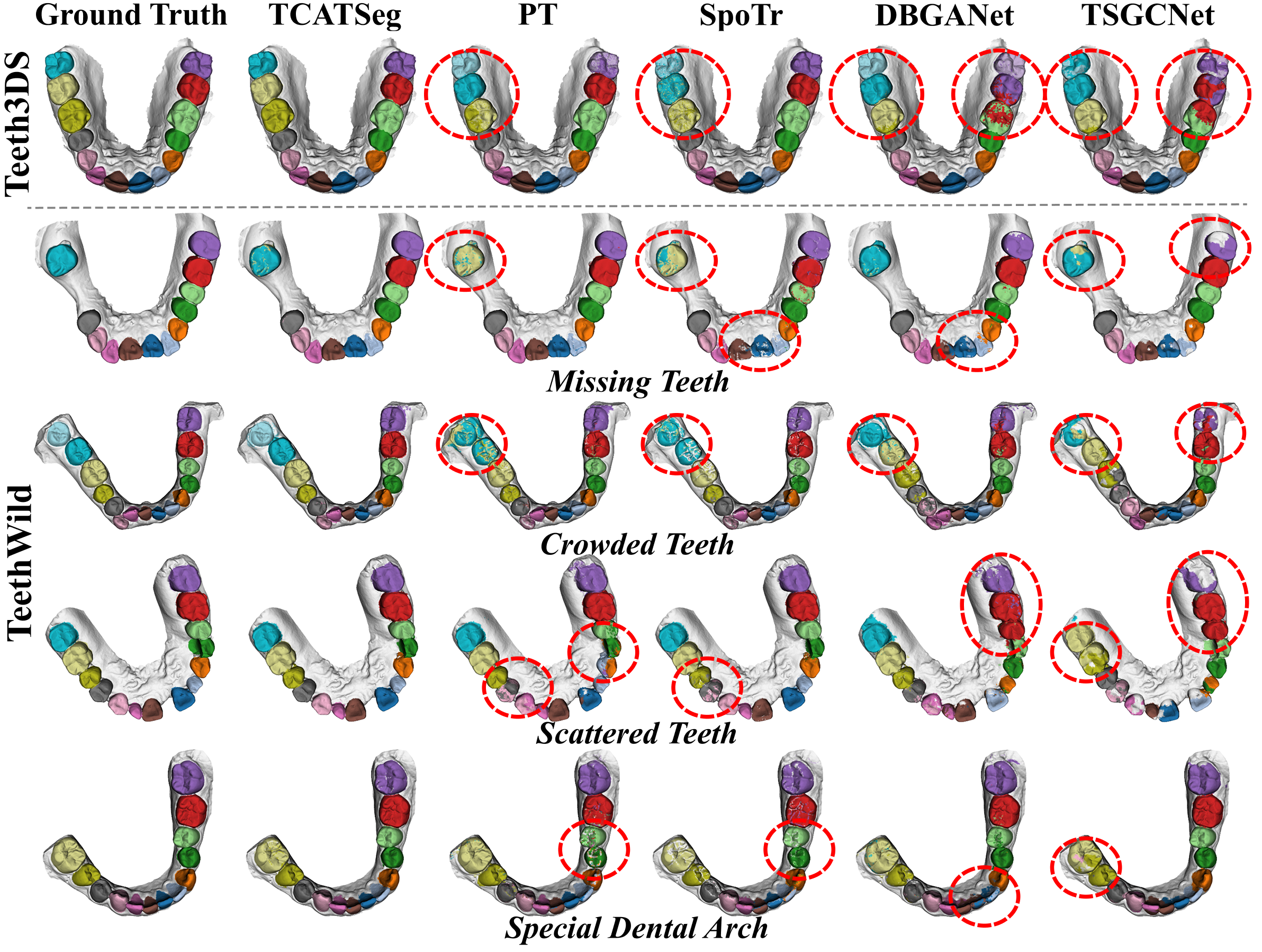} 

\caption{Tooth segmentation results on Teeth3DS and TeethWild, with errors highlighted by red dashed lines.}

\label{fig-vis}
\end{figure}

We perform a comprehensive comparison between our TCATSeg and several state-of-the-art baselines across three categories:
1) classical methods, including PointNet++~\cite{qi2017pointnet++}, DGCNN~\cite{wang2019dynamic}, PointTransformer(PT)~\cite{zhao2021point} and TSGCNet~\cite{zhang2021tsgcnet};
2) methods designed for modeling global contexts, such as SpoTr~\cite{park2023self} and DBGANet~\cite{lin2023dbganet};
3) domain-specific methods from 3DTeethSeg'22 challenge~\cite{ben20233dteethseg}.

\begin{figure}[!t]
  \centering
  \includegraphics[width=0.7\linewidth]{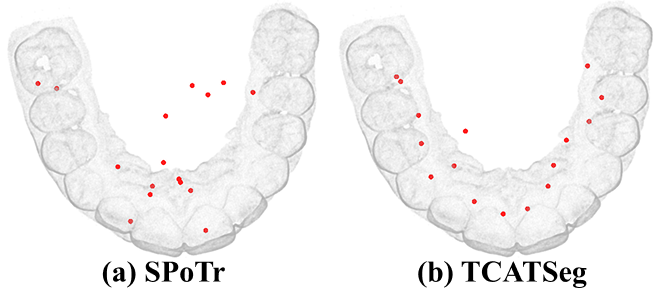}
  \caption{The comparison of superpoints in SpoTr and TCATSeg. Red points correspond to the superpoints.}
  \label{fig:hyperpoints}
\end{figure}

We evaluate these methods on the Teeth3DS dataset using patient-based splits (Table~\ref{tab:experiment1}).
TCATSeg consistently outperforms all baselines in all metrics, demonstrating superior segmentation accuracy.
In particular, we compare methods designed for modeling global context with TCATSeg.
Although DBGANet leverages a grouped global attention mechanism to model long-range dependencies,
and SpoTr adopts self-positioning to capture global information,
both methods show worse TIR compared to TCATSeg.
To assess the role of superpoints, we visualize their distributions in Fig.~\ref{fig:hyperpoints}. SpoTr's superpoints are chaotically distributed, whereas TCATSeg's align closely with tooth arrangement, indicating more effective semantic context modeling.

\vspace{-2mm}

\begin{table}[ht]
  \centering
  \fontsize{9pt}{11pt}\selectfont
  
  \caption{Segmentation results on Teeth3DS and TeethWild.}
  \label{tab:experiment1}
  {
    \begin{tabular}{l|l|cccccc}
    \toprule
    D. & Method & TIR$\uparrow$ & SEN$\uparrow$ & DSC$\uparrow$ & PPV$\uparrow$ & OA$\uparrow$ \\
    \midrule
    \multirow{7}{*}{\rotatebox{270}{Teeth3DS}}
    & TSGCNet          & 95.17 & 92.32 & 89.64 & 91.26 & 94.67 \\
    & DBGANet          & 95.91 & 94.30 & 89.82 & 90.65 & 95.54 \\
    & DGCNN            & 95.77 & 91.73 & 89.94 & 92.23 & 94.35 \\
    & PointNet++       & 95.31 & 93.17 & 90.87 & 92.40 & 95.04 \\
    & PT               & 95.80 & 95.29 & 93.51 & 94.58 & 96.30 \\
    & SPoTr            & 95.92 & 95.13 & 93.45 & 94.49 & 96.15 \\
    \cmidrule{2-7}
    & \textbf{TCATSeg} & \textbf{96.48} & \textbf{95.51} & \textbf{93.92} & \textbf{94.65} & \textbf{96.36} \\
    \midrule
    \multirow{7}{*}{\rotatebox{270}{TeethWild}}
    & TSGCNet          & 77.91 & 65.25 & 60.31 & 66.30 & 77.35 \\
    & DBGANet          & 88.64 & 79.79 & 68.79 & 69.79 & 84.57 \\
    & DGCNN            & 89.17 & 74.01 & 70.90 & 80.09 & 83.87 \\
    & PointNet++       & 85.31 & 77.66 & 70.00 & 74.43 & 84.47 \\
    & PT               & 88.46 & 86.72 & 78.93 & 80.51 & 88.59 \\
    & SPoTr            & 89.77 & 86.43 & 80.77 & 82.86 & 89.93 \\
    \cmidrule{2-7}
    & \textbf{TCATSeg} & \textbf{89.99} & \textbf{88.86} & \textbf{82.27} & \textbf{83.34} & \textbf{90.26} \\
    \bottomrule
    \end{tabular}
  }
\end{table}
\vspace{-2mm}

\begin{table}[ht]
  \centering
  \caption{The results of 3DTeethSeg'22 Challenge~\cite{ben20233dteethseg}.}
  \label{tab:experiment2}
  \fontsize{9pt}{11pt}\selectfont
  {
    \begin{tabular}{l|cccc}
    \midrule 
    \fontsize{9}{10}\selectfont
    Method & TLA$\uparrow$ & TSA$\uparrow$ & TIR$\uparrow$ & Score$\uparrow$ \\
    \midrule
    CGIP    & 96.58 & \textbf{98.59} & 91.00 & 95.39 \\
    FiboSeg & \textbf{99.24} & 92.93 & 92.23 & 94.80 \\
    IGIP    & 92.44 & 97.50 & 92.89 & 94.27 \\
    TeethSeg & 91.84 & 96.78 & 85.38 & 91.33 \\
    OS      & 78.45 & 96.93 & 89.40 & 88.26 \\
    Chompers & 62.42 & 88.86 & 87.95 & 79.74 \\
    \midrule
    \textbf{TCATSeg} & 98.53 & 96.54 & \textbf{95.48} & \textbf{96.85} \\
    \midrule
    \end{tabular}
  }
\end{table}

\vspace{-2mm}
To evaluate cross-domain generalization, 
we perform zero-shot evaluation by directly applying models trained on Teeth3DS patient-wise splits to the entire TeethWild dataset, without any fine-tuning.
As shown in Table~\ref{tab:experiment1}, although the performance of all methods decreased,
TCATSeg still outperforms these baselines in all metrics, demonstrating its superior generalization.
We illustrate the visualization of semantic tooth segmentation results in Fig.~\ref{fig-vis}. 
Notably, even when confronted with challenging samples in TeethWild, such as spatially scattered or crowded teeth, TCATSeg consistently achieves accurate segmentation.

Additionally, we utilize the official dataset partitioning and evaluation code from the 3DTeethSeg'22 challenge to further assess the effectiveness of our method. As shown in Table \ref{tab:experiment2}, the overall score achieved by TCATSeg placed our method in the leading position on the challenge leaderboard. Notably, the Teeth Identification Rate significantly exceeds the existing methods, further demonstrating TCATSeg's ability to effectively model global semantic information.
\vspace{-2mm}
\begin{table}[ht]
\caption{Ablation of key components in SGDA and DPDA.}
\centering
\fontsize{9pt}{11pt}\selectfont 
\begin{adjustbox}{width=\linewidth,keepaspectratio}
\begin{tabular}{l|lllll|cccccc}
\midrule 
No. &LocA & SG & GA & LayA & $\mathcal{L}_{\text{offset}}$ & TIR$\uparrow$ & SEN$\uparrow$ & DSC$\uparrow$  & OA$\uparrow$ \\
\midrule
1 & \checkmark & ~ & ~  & ~ & ~ &  96.06 & 95.11 & 93.70  & 96.21 \\
2 & \checkmark & \checkmark & ~ & ~ & ~ &  95.85 & 94.93 & 93.22  & 96.08  \\
3 & \checkmark & \checkmark & \checkmark  & ~ & ~ & 96.33 & 95.36 & 93.82 & 96.32\\
4 & \checkmark & \checkmark & \checkmark  & \checkmark & ~ & 96.46 & 95.45 & 93.81  &96.34\\
5 & \checkmark & \checkmark & \checkmark  & \checkmark & \checkmark & \textbf{96.48} & \textbf{95.51} & \textbf{93.92}  &\textbf{96.36}\\
\midrule
\end{tabular}
\end{adjustbox}
\label{ablation}
\end{table}

\vspace{-7mm}
\subsection{Ablation Study}
We evaluated the contribution of each component in TCATSeg through ablation studies (Table~\ref{ablation}). Using only local attention (No.1) yields limited performance, while incorporating superpoints with $\mathcal{L}_{tcp}$ (No.2) without further optimization remains suboptimal. Optimizing superpoints via DPDA (No.3--5) enhances performance by leveraging global context. Specifically, GA (No.3 vs. No.2) improves all metrics through effective semantic aggregation, while LayA (No.4 vs. No.3) further boosts TIR by enforcing layer-wise consistency of superpoint representations. Finally, introducing the offset loss (No.5) refines the offset head and achieves the best overall performance, ensuring compatibility with two-stage dental segmentation frameworks~\cite{ben20233dteethseg,cui2021tsegnet}.
\vspace{-2mm}
\section{Conclusion}
\vspace{-2mm}
We propose TCATSeg, a novel superpoint-guided network for tooth semantic segmentation.
By constraining superpoint feature distribution close to the teeth centroids and integrating global semantic features into local geometric features, the proposed method could aggregate the semantic arrangement and local geometry information of teeth effectively. 
Experiments on our collected challenging dataset and public benchmark show that TCATSeg outperforms state-of-the-art methods.
In the future, we will further improve the robustness of our method for real-world dental applications.

\section{ACKNOWLEDGEMENTS}

This work was supported by the National Natural Science Foundation of China (Grant Nos. 62272447, 62502337) and the Beijing Natural Science Foundation Haidian Original Innovation Joint Fund (Grant Nos. L222008, L232028, and L242060).

\bibliographystyle{IEEEbib}
\bibliography{strings,refs}

\end{document}